\documentclass[sigconf]{acmart}

\usepackage{makecell}
\usepackage{booktabs}
\usepackage{multirow}
\usepackage{xcolor}
\usepackage{colortbl}

\usepackage{amsmath}
\usepackage{amssymb}
\usepackage{amsfonts}
\usepackage{graphicx}
\usepackage{algorithm}
\usepackage{algorithmic}
\usepackage{etoolbox}

\AtBeginDocument{%
  }

\copyrightyear{2026}
\acmYear{2026}
\setcopyright{cc}
\setcctype{by}
\acmConference[KDD '26]{Proceedings of the 32nd ACM SIGKDD Conference on Knowledge Discovery and Data Mining V.2}{August 09--13, 2026}{Jeju Island, Republic of Korea}
\acmBooktitle{Proceedings of the 32nd ACM SIGKDD Conference on Knowledge Discovery and Data Mining V.2 (KDD '26), August 09--13, 2026, Jeju Island, Republic of Korea}
\acmISBN{979-8-4007-2259-2/2026/08}
\acmDOI{10.1145/3770855.3817786}
\title{AnchorMoE: Interpretable Time Series Classification via Anchor-Routed MoE}

\author{Tao Xie}
\affiliation{
  \institution{Guangdong University of Technology}
  \city{Guangzhou}
  \country{China}
}
\email{3123001085@mail2.gdut.edu.cn}

\author{Zexi Tan}
\affiliation{
  \institution{Guangdong University of Technology}
  \city{Guangzhou}
  \country{China}
}
\email{3123004194@mail2.gdut.edu.cn}

\author{Haoyi Xiao}
\affiliation{
  \institution{Guangdong University of Technology}
  \city{Guangzhou}
  \country{China}
}
\email{3124004599@mail2.gdut.edu.cn}

\author{Mengke Li}
\affiliation{
  \institution{Shenzhen University}
  \city{Shenzhen}
  \country{China}
}
\email{mengkeli@szu.edu.cn}

\author{Yiqun Zhang}
\authornote{Corresponding author.}
\affiliation{
  \institution{Guangdong University of Technology}
  \city{Guangzhou}
  \country{China}
}
\email{yqzhang@gdut.edu.cn}

\author{Yang Lu}
\affiliation{
  \institution{Xiamen University}
  \city{Xiamen}
  \country{China}
}
\email{luyang@xmu.edu.cn}

\author{Cuie Yang}
\affiliation{
  \institution{Northeastern University}
  \city{Shenyang}
  \country{China}
}
\email{yangcuie@mail.neu.edu.cn}

\author{Yiu-ming Cheung}
\affiliation{
  \institution{Hong Kong Baptist University}
  \city{Hong Kong SAR}
  \country{China}
}
\email{ymc@comp.hkbu.edu.hk}

\renewcommand{\shortauthors}{Tao Xie et al.}

\begin{document}



\begin{abstract}

Multivariate time series classification (MTSC) is pivotal in high-stakes domains, such as clinical diagnosis and industrial fault detection, where safe deployment necessitates transparent decision-making. However, isolating the temporal segments that drive model predictions is challenging because discriminative signals in real-world time series are typically sparse, heterogeneous, and heavily obscured by background noise. This paper, therefore, proposes AnchorMoE, an interpretable-by-construction classification framework. Built upon a Mixture-of-Experts (MoE) architecture, AnchorMoE encodes multi-view representations of local patches and routes them to specialized experts, ensuring that the final prediction is formulated as an exact additive decomposition over the input segments, facilitating ante-hoc transparency rather than relying on post-hoc estimations. To maintain the reliability of this decomposition under sparse signal distributions, we introduce a geometric orthogonality constraint that penalizes representational redundancy, compelling distinct experts to specialize in heterogeneous predictive patterns. Furthermore, an uncertainty-aware reliability gate is designed to dynamically calibrate the contribution of each segment, effectively suppressing residual background noise. Extensive experiments on real-world and synthetic benchmarks demonstrate that AnchorMoE achieves highly competitive classification performance while faithfully grounding its decisions in the raw time series.
\end{abstract}

\maketitle
\newcommand\kddavailabilityurl{https://doi.org/10.5281/zenodo.20449598}
\ifdefempty{\kddavailabilityurl}{}{
\begingroup\small\noindent\raggedright\textbf{Resource Availability:}\\
The source code of this paper has been made publicly available at \url{\kddavailabilityurl} (GitHub: \url{https://github.com/kuxit/AnchorMoE}).
\endgroup
}

\section{Introduction}

Multivariate Time Series (MTS) capture complex, co-evolving measurements pervasive in domains like network security, industrial monitoring~\cite{lian2026decompose, lian2026contextual}, and clinical care~\cite{xie2025de3s, tan2025meetsepsis}. 
In these high-stakes applications, MTS Classification (MTSC) is crucial for state identification, such as detecting equipment failure or predicting patient deterioration. Because these predictions dictate critical interventions, accuracy must be accompanied by transparency, thus the specific temporal segments driving the model's decision should be isolated. However, localizing this evidence is fundamentally challenging. In real-world MTS, class-discriminative signals are typically sparse, localized within brief temporal windows, and restricted to a subset of variables, while the remainder of the sequence constitutes uninformative background. Furthermore, this evidence is highly heterogeneous, comprising distinct patterns across different variables, timeframes, and frequency domains~\cite{zhang2023learning} that jointly determine the class label. This structural sparsity and diversity are well-documented in the literature~\cite{chen2024timemil, tan2026mask}, 
and the multivariate nature of the data further exacerbates the difficulty by distributing the signal unevenly across spatial and temporal dimensions~\cite{hsieh2021explainable, dhariyal2021fast}.

\begin{figure}[t!]
  \centering
  \includegraphics[width=\linewidth]{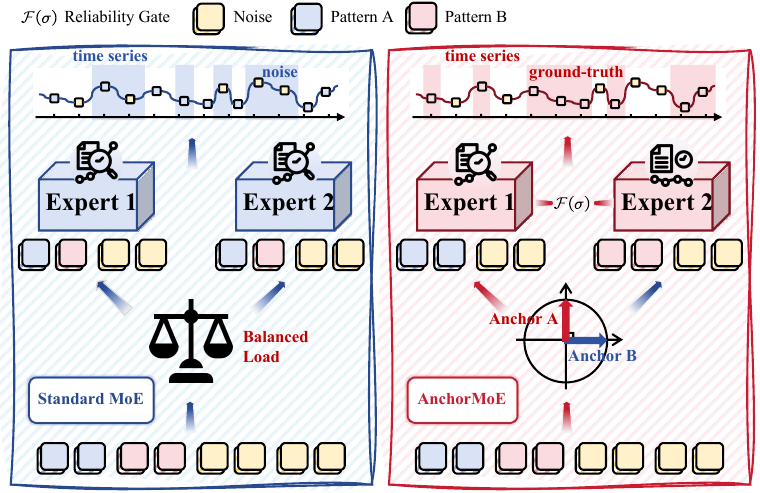}
  \caption{Standard MoE (left) vs. AnchorMoE (right). Conventional MoE enforces uniform routing, causing experts to learn redundant representations and absorb background noise from sparse MTSC signals. To resolve this, AnchorMoE employs orthogonal posterior anchors for distinct expert specialization, while an uncertainty-aware reliability gate dynamically suppresses uninformative segments, ensuring highly localized and faithful attribution.}
  \label{fig:motivation}
\end{figure}

To address this, the prevailing paradigm relies on post-hoc attribution, which estimates the contribution of input regions after prediction. Existing techniques span multiple families, including gradient-based methods (e.g., Integrated Gradients
~\cite{sundararajan2017axiomatic} and DeepLIFT~\cite{shrikumar2017deeplift}), perturbation-based masking (e.g., DynaMask~\cite{crabbe2021dynamask} and FIT~\cite{tonekaboni2020fit}), surrogate self-explainers (e.g., TimeX~\cite{queen2023timex}), and 
the heuristic interpretation of attention weights.
However, 
each approach presents specific vulnerabilities: gradients suffer from saturation~\cite{sundararajan2017axiomatic}, 
perturbations often push inputs off the data manifold, surrogates lack guaranteed fidelity to the target model, and attention weights reflect information routing rather than causal feature importance~\cite{jain2019attention}. More fundamentally, 
these methods share a structural limitation, i.e., the explanation process is decoupled from the predictive computation. Accordingly, since attribution and decision-making are driven separately, the faithfulness of the explanation requires external verification, which itself may be unreliable~\cite{adebayo2018sanity, turbe2023evaluation}. 
Given the sparse and scattered nature of the signal, the above-mentioned post-hoc methods can easily assign high attribution to correlated background noise, with no architectural safeguard to prevent the mis-attribution. 

A promising alternative is to construct models that are interpretable by design, often referred to as ante-hoc or interpretable-by-construction architectures~\cite{rudin2019stop, di2025ante}. The most transparent structure for such models is additive, where the final prediction is formulated as an explicit sum of contributions from individual input segments, allowing direct inspection of each segment's role. The Mixture-of-Experts (MoE) routing~\cite{shazeer2017outrageously} naturally facilitates this decomposition. By employing a gating function to dispatch segments to specialized experts, the network constrains each segment to contribute solely through its assigned expert, rendering the prediction an exact summation of per-segment terms. Consequently, classification and explanation unify into a single process, facilitating a faithful decomposition without relying on external attributors. 

Such architectural faithfulness, however, solely ensures that the explanation reflects internal computational logic, and does not guarantee that the model relies on genuinely discriminative segments. That is, a faithful but poorly grounded model will accurately report its reliance on background noise, which is still incompetent in interpreting accurate MTSC. In other words, standard routing mechanisms fail to resolve the key problem because conventional MoE architectures drive expert specialization through load-balancing objectives that enforce uniform utilization~\cite{shazeer2017outrageously, lepikhin2021gshard, fedus2022switch}. This is structurally misaligned with MTSC, as distributing predominantly uninformative sequence data evenly across experts compels them to absorb background noise rather than isolating distinct and predictive patterns. Consequently, experts learn overlapping representations, allowing uninformative background noise to enter the additive composition as genuine evidence, ultimately yielding diffuse and unreliable attributions (Figure~\ref{fig:motivation}, left). 

This paper, therefore, proposes \textbf{AnchorMoE}, an ante-hoc interpretable classification framework. Within this framework, each multivariate patch is first enriched into a comprehensive evidence unit integrating temporal, spectral, and contextual representations. These units are then routed to specialized experts, formulating the final prediction as a transparent, inspectable sum of per-patch contributions. To ensure this decomposition remains reliable under signal pattern heterogeneity, we extract a posterior anchor characterizing the evidence processed by each expert. By enforcing mutual orthogonality among these anchors, distinct experts are compelled to specialize in heterogeneous patterns, preventing representational smoothing. Furthermore, to prevent uninformative background segments from artificially inflating the sum, each segment's contribution is dynamically scaled via an uncertainty-aware reliability gate prior to composition. Extensive evaluations across real-world and synthetic benchmarks demonstrate that AnchorMoE is extremely competitive in classification performance against strong baselines, while explicitly grounding its decisions in the raw data (Figure~\ref{fig:motivation}, right). 
The main contributions are: 
\begin{itemize}
    \item 
    To overcome the inherent unreliability of decoupled post-hoc explanations in MTSC, we formulate the MTSC problem as an interpretable-by-design expert-routing problem. This ensures attribution as an exact, additive decomposition of the decision process rather than a post-hoc approximation. 
    \item 
    AnchorMoE is introduced with orthogonal posterior anchors and uncertainty-aware reliability gate to prevent the additive decomposition from being corrupted by sparse signals and background noise. Such design structurally compels experts to specialize in heterogeneous predictive patterns and dynamically suppress uninformative segments.
    \item 
    Rigorous evaluation of AnchorMoE is designed to demonstrate that strict ante-hoc transparency does not necessitate a compromise in predictive performance. To substantiate this, datasets are constructed with controlled distractors to confirm that AnchorMoE rivals state-of-the-art classifiers in accuracy while yielding uniquely faithful decompositions. 
\end{itemize}

\section{Related Work}
This section reviews related work in three key areas: post-hoc explanation, interpretable time series models, and MoE architectures.

\subsection{Post-hoc Explanation}
Post-hoc explanation methods seek to interpret trained models by attributing predictions to salient input features. These approaches can be broadly categorized by their underlying mechanisms. Gradient-based techniques (e.g., Grad-CAM~\cite{selvaraju2017gradcam}, Integrated Gradients~\cite{sundararajan2017axiomatic}, DeepLIFT~\cite{shrikumar2017deeplift}) quantify feature relevance by backpropagating prediction signals through the network. Perturbation-based methods instead estimate importance by measuring output variations under input alterations, including model-agnostic frameworks (e.g., LIME~\cite{ribeiro2016lime}, SHAP~\cite{lundberg2017shap}) and time-series-specific masking strategies (e.g., DynaMask~\cite{crabbe2021dynamask}, FIT~\cite{tonekaboni2020fit}). Other methods train surrogate models (e.g., TimeX~\cite{queen2023timex}) or extract attention weights as attributions, although the theoretical validity of attention-based attribution remains contested~\cite{jain2019attention}. The aforementioned methods decouple explanation generation from the underlying predictive computation, necessitating unreliable external faithfulness validation~\cite{adebayo2018sanity, turbe2023evaluation}. This limitation has motivated a transition toward interpretable-by-construction models~\cite{rudin2019stop, di2025ante}, where explanatory mechanisms are embedded directly into the forward inference process. 

\subsection{Interpretable Time Series Models}
Inherently interpretable time series models originated with shapelets \cite{ye2009shapelets, grabocka2014learning}, which provide explanatory signals by aligning discriminative subsequences with inputs. Parallel research characterizes classes using learned prototypes~\cite{chen2019looks, ming2019prototypes}, a paradigm later adapted to time series~\cite{ghods2022pip}. Recent architectures focus on the contributions of local temporal units. Specifically, SoftShape~\cite{liu2025softshape} routes learned soft shapelets along class-specific paths. Concurrently, InterpGN~\cite{wen2025interpgn} uses a gating mechanism to integrate an interpretable shapelet branch with a deep representation model. Beyond the above advances, other frameworks achieve local interpretability by attending to critical variables and temporal intervals~\cite{hsieh2021explainable} or by using masking-guided attention to isolate salient features~\cite{zhang2023time}. Nevertheless, extending these inherently interpretable designs to multivariate settings remains underexplored.

\subsection{Mixture-of-Experts}
Routing local representations to specialized computational units is the foundational mechanism of MoE architectures~\cite{shazeer2017outrageously, fedus2022switch, zhou2022expertchoice}. A primary challenge in scaling these models is the frequent failure of experts to achieve intended specialization, which often culminates in representation collapse~\cite{chi2022collapse}. Recent work mitigates this degradation by enforcing orthogonality constraints across expert modules, either within low-rank adapters~\cite{feng2025omoe} or across heterogeneous objectives in multi-task learning~\cite{hendawy2024moore}. In the time series domain, MoE architectures are predominantly deployed for forecasting. Large-scale approaches, such as Time-MoE~\cite{shi2024timemoe} and Moirai-MoE~\cite{liu2024moiraimoe}, integrate sparse experts into foundation models to handle scale. Conversely, smaller-scale frameworks like MoLE~\cite{ni2024mixture} and FMoE~\cite{chen2026fmoe} employ lightweight linear and frequency-guided experts, respectively. The most closely related work~\cite{ismail2023interpretable} attains inspectability by exposing explicit expert assignments. 
However, its reliance on standard expert networks limits its capacity to capture the complex inter-variable dependencies and spatiotemporal heterogeneity essential to MTSC, the focus of this work.

\begin{figure*}[t!]
  \centering
  \includegraphics[width=0.92\linewidth]{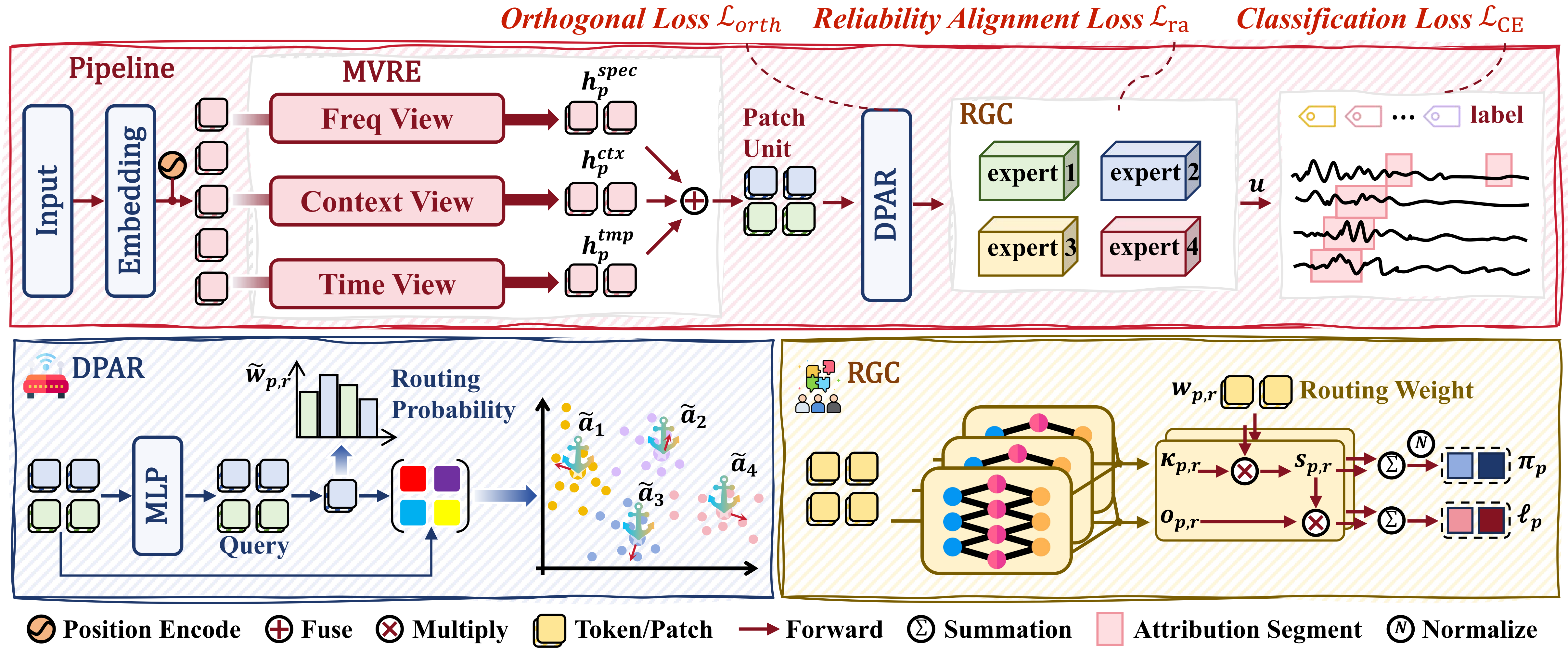}
  \caption{Overview of AnchorMoE. Multi-View Representation Embedding (MVRE) encodes temporal, spectral, and contextual patch cues. Diversified Posterior-Anchor Routing (DPAR) routes patches while regularizing anchors to prevent expert overlap. Reliability-Gated Composition (RGC) aggregates calibrated logits into an explicitly decomposable additive prediction.}
  \label{fig:framework}
\end{figure*}

\section{Proposed Method}
This section first formulates the problem, and then presents the three components of AnchorMoE: Multi-View Representation Embedding (MVRE) encodes each patch, Diversified Posterior-Anchor Routing (DPAR) specializes the experts, and Reliability-Gated Composition (RGC) calibrates each contribution, which are finally combined into an additive prediction. The overall architecture is summarized in Figure~\ref{fig:framework}.

\subsection{Problem Formulation}
\label{sec:PF}
Consider a multivariate time series sample $\boldsymbol{X}\in\mathbb{R}^{\mathrm{D}\times\mathrm{T}}$, where $\mathrm{D}$ and $\mathrm{T}$ denote the numbers of variables and time steps. The corresponding label $y$ resides in an $\mathrm{n}_{\mathrm{c}}$-class label space. MTSC learns a classifier mapping $\boldsymbol{X}$ to a logit vector $\boldsymbol{u}\in\mathbb{R}^{\mathrm{n}_{\mathrm{c}}}$, with the predicted class $\hat{y}=\operatorname{argmax}_{\mathrm{c}}u_{\mathrm{c}}$. 
For interpretable MTSC, prediction accuracy alone is insufficient. Each class logit should be admitted with a patch-level decomposition to ensure that local terms supporting $\hat{y}$ are directly inspectable.

To define these local units, $\boldsymbol{X}$ is divided along the time axis into $\mathrm{P}$ multivariate patches. Given patch length $\mathrm{L}$ and stride $\mathrm{s}$, the $\mathrm{p}$-th patch is $\boldsymbol{x}_{\mathrm{p}}\in\mathbb{R}^{\mathrm{D}\times\mathrm{L}}$. A binary validity mask $m_{\mathrm{p}}$ tracks observed time steps, excluding padded windows from routing and composition. The patch set of $\boldsymbol{X}$ is denoted by:
\begin{equation}
\mathcal{X}
=
\left\{
\left(
\boldsymbol{x}_{\mathrm{p}},
m_{\mathrm{p}}
\right)
\right\}_{\mathrm{p}=1}^{\mathrm{P}} .
\label{eq_patch_set}
\end{equation}
AnchorMoE requires each class logit to be decomposed over valid patches. In particular, the target form is:
\begin{equation}
u_{\mathrm{c}} = \sum_{\mathrm{p}=1}^{\mathrm{P}} A_{\mathrm{p},\mathrm{c}},
\label{eq_decomposition_goal}
\end{equation}
where $A_{\mathrm{p},\mathrm{c}}$ denotes the contribution of the $\mathrm{p}$-th patch to the $\mathrm{c}$-th class, with $A_{\mathrm{p},\mathrm{c}}=0$ if $m_{\mathrm{p}} = 0$. Such an additive structure resolves the decoupling inherent in post-hoc explanations, as it ensures that patch-level contributions are readable directly from the forward pass. Accordingly, attribution becomes inherently homologous to the predictive computation, obviating the need for external, unreliable post-hoc estimators. Thus, $\mathcal{X}$ serves as the unified foundation for the following patch representation, expert assignment, and additive score construction.

\subsection{Multi-View Representation Embedding}
\label{sec:MVRE}

Raw local patches lack sufficient discriminative information, while class relevance typically depends on a complex combination of temporal patterns, frequency content, and sequence-level context. To tackle this, MVRE encodes these multi-view features into holistic semantic space to provide robust and information-rich inputs for the downstream expert routing. 

Given the $\mathrm{p}$-th patch $\boldsymbol{x}_{\mathrm{p}}$, a channel-joint embedding layer maps it into a token with positional encoding:
\begin{equation}
\boldsymbol{h}_{\mathrm{p}}
=
\psi_{\mathrm{emb}}
\left(
\boldsymbol{x}_{\mathrm{p}}
\right)
\in
\mathbb{R}^{\mathrm{d}} .
\label{eq_mvre_embed}
\end{equation}
Here $\psi_{\mathrm{emb}}$ denotes a convolutional embedding function followed by normalization and positional encoding. The temporal view preserves ordered local morphology through:
\begin{equation}
\boldsymbol{h}_{\mathrm{p}}^{\mathrm{tmp}}
=
\boldsymbol{h}_{\mathrm{p}}
+
g_{\mathrm{tmp}}
\psi_{\mathrm{t}}
\left(
\boldsymbol{h}_{\mathrm{p}}
\right),
\label{eq_mvre_temporal}
\end{equation}
where $g_{\mathrm{tmp}}=\sigma\left(\theta_{\mathrm{tmp}}\right)$ is a learnable gate and $\theta_{\mathrm{tmp}}$ is a scalar parameter.

To complement the time-domain cue, MVRE summarizes frequency content through $\mathrm{B}$ frequency bands. For the $\mathrm{j}$-th variable and the $\mathrm{b}$-th band, the energy is computed as:
\begin{equation}
E_{\mathrm{p},\mathrm{j},\mathrm{b}}
=
\sum_{\mathrm{f}\in\mathcal{B}_{\mathrm{b}}}
\left|
\mathcal{F}
\left(
\boldsymbol{x}_{\mathrm{p},\mathrm{j}}
\right)_{\mathrm{f}}
\right|^{2},
\label{eq_mvre_energy}
\end{equation}
where $\mathcal{B}_{\mathrm{b}}$ and $\mathcal{F}$ denote the $\mathrm{b}$-th band and Fourier transform, respectively. Let $\boldsymbol{v}_{\mathrm{p}}$ collect the log-scaled band energies over all variables and bands:
\begin{equation}
\boldsymbol{v}_{\mathrm{p}}=\operatorname{vec}\left(\left\{\log\left(1+E_{\mathrm{p},\mathrm{j},\mathrm{b}}\right)\right\}_{\mathrm{j}=1,\mathrm{b}=1}^{\mathrm{D},\mathrm{B}}\right),
\label{eq_mvre_vec}
\end{equation}
then the spectral view can be written as:
\begin{equation}
\boldsymbol{h}_{\mathrm{p}}^{\mathrm{spec}}
=
\psi_{\mathrm{f}}
\left(
\boldsymbol{v}_{\mathrm{p}}
\right),
\label{eq_mvre_spectral}
\end{equation}
which captures frequency-level regularities that are less dependent on exact temporal alignment.

MVRE further introduces a contextual view that situates each patch within the whole series, so that a patch is encoded not in isolation but relative to the global context. To this end, MVRE first summarizes all valid patches into a reference centroid, then scores each token against this centroid, and finally lifts the score into a contextual representation:
\begin{equation}
\bar{\boldsymbol{h}}
=
\frac{\sum_{\mathrm{i}=1}^{\mathrm{P}} m_{\mathrm{i}} \boldsymbol{h}_{\mathrm{i}}}
     {\sum_{\mathrm{i}=1}^{\mathrm{P}} m_{\mathrm{i}}},
\qquad
r_{\mathrm{p}}
=
\frac{\left\langle \boldsymbol{h}_{\mathrm{p}}, \bar{\boldsymbol{h}} \right\rangle}
     {\sqrt{\mathrm{d}}},
\qquad
\boldsymbol{h}_{\mathrm{p}}^{\mathrm{ctx}}
=
\psi_{\mathrm{ctx}}\left( r_{\mathrm{p}} \right).
\label{eq_mvre_context}
\end{equation}
Here $\bar{\boldsymbol{h}}$ is the mask-weighted centroid of all valid patch tokens, which serves as a sequence-level reference and excludes padded windows through $m_{\mathrm{i}}$. The scalar $r_{\mathrm{p}}$ is the scaled alignment between the $\mathrm{p}$-th token and this reference, where the division by $\sqrt{\mathrm{d}}$ stabilizes the inner product across dimensions. Finally, $\psi_{\mathrm{ctx}}$ maps the relation score $r_{\mathrm{p}}$ to a contextual embedding $\boldsymbol{h}_{\mathrm{p}}^{\mathrm{ctx}}\in\mathbb{R}^{\mathrm{d}}$, giving each local patch a sequence-aware reference before expert routing.

The three views $\boldsymbol{h}_{\mathrm{p}}^{\mathrm{tmp}}$, $\boldsymbol{h}_{\mathrm{p}}^{\mathrm{spec}}$, and $\boldsymbol{h}_{\mathrm{p}}^{\mathrm{ctx}}$ extracted by Eqs.~(\ref{eq_mvre_temporal}), (\ref{eq_mvre_spectral}), and (\ref{eq_mvre_context}), respectively, are ultimately fused into the patch unit $\boldsymbol{e}_{\mathrm{p}}$:
\begin{equation}
\boldsymbol{e}_{\mathrm{p}}
=
\operatorname{LN}
\left(
\boldsymbol{h}_{\mathrm{p}}^{\mathrm{tmp}}
+
g_{\mathrm{spec}}\,\boldsymbol{h}_{\mathrm{p}}^{\mathrm{spec}}
+
g_{\mathrm{ctx}}\,\boldsymbol{h}_{\mathrm{p}}^{\mathrm{ctx}}
\right),
\label{eq_mvre_fusion}
\end{equation}
where $g_{\mathrm{spec}}=\sigma\left(\theta_{\mathrm{spec}}\right)$ and $g_{\mathrm{ctx}}=\sigma\left(\theta_{\mathrm{ctx}}\right)$ are learnable gates that balance the spectral and contextual cues against the temporal backbone. The fused unit $\boldsymbol{e}_{\mathrm{p}}$ then produces the routing query $\boldsymbol{q}_{\mathrm{p}}=\psi_{\mathrm{q}}\left(\boldsymbol{e}_{\mathrm{p}}\right)$ and is also passed to the experts for score composition. So far, the shared representation that will be further processed by DPAR and RGC has been facilitated.

\subsection{Diversified Posterior-Anchor Routing}
To prevent experts from converging on homogeneous patterns, DPAR regularizes the geometric diversity of assigned representations. Given the routing query $\boldsymbol{q}_{\mathrm{p}}$ obtained through Eq.~(\ref{eq_mvre_fusion}) in Section~\ref{sec:MVRE}, the probability $\tilde{w}_{\mathrm{p},\mathrm{r}}$ of assigning the $\mathrm{p}$-th patch to the $\mathrm{r}$-th expert is computed via a learnable router $\psi_{\mathrm{g}}$:
\begin{equation}
\tilde{w}_{\mathrm{p},\mathrm{r}}
=
\operatorname{softmax}_{\mathrm{r}}
\left(
\psi_{\mathrm{g}}
\left(
\boldsymbol{q}_{\mathrm{p}}
\right)
\right),
\label{eq_dpar_routing}
\end{equation}
where $\operatorname{softmax}_{\mathrm{r}}$ normalizes the output into a probability distribution over the $\mathrm{M}$ experts.

To capture which patch representations are routed to each expert, DPAR summarizes them into a single representative vector per expert. Concretely, for the $\mathrm{r}$-th expert, the patch units $\boldsymbol{e}_{\mathrm{p}}$ are averaged with their routing probabilities $\tilde{w}_{\mathrm{p},\mathrm{r}}$ as weights, yielding the posterior anchor $\tilde{\boldsymbol{a}}_{\mathrm{r}}$:
\begin{equation}
\tilde{\boldsymbol{a}}_{\mathrm{r}}
=
\frac{
\sum_{\mathrm{p}=1}^{\mathrm{P}}
m_{\mathrm{p}}
\tilde{w}_{\mathrm{p},\mathrm{r}}
\boldsymbol{e}_{\mathrm{p}}
}{
\sum_{\mathrm{p}=1}^{\mathrm{P}}
m_{\mathrm{p}}
\tilde{w}_{\mathrm{p},\mathrm{r}}
+
\epsilon
},
\label{eq_dpar_anchor}
\end{equation}
where the validity mask $m_{\mathrm{p}}$ excludes padded patches from the average, and $\epsilon$ is a small constant that prevents division by zero when an expert receives negligible routing weight. Unlike statically predefined parameters, $\tilde{\boldsymbol{a}}_{\mathrm{r}}$ acts as a \emph{posterior} anchor dynamically induced by the routing distribution $\tilde{w}_{\mathrm{p},\mathrm{r}}$ in Eq.~(\ref{eq_dpar_routing}). Geometrically, it represents the weighted centroid of the patch units allocated to the $\mathrm{r}$-th expert, establishing the basis for the subsequent diversification.

Given these posterior anchors, expert redundancy is penalized via an orthogonal loss based on their pairwise cosine similarity:
\begin{equation} 
\mathcal{L}_{\mathrm{orth}} = \frac{1}{\mathrm{M}\left(\mathrm{M}-1\right)} \sum_{\mathrm{r}\ne \mathrm{s}} \left( \frac{ \left\langle \tilde{\boldsymbol{a}}_{\mathrm{r}}, \tilde{\boldsymbol{a}}_{\mathrm{s}} \right\rangle }{ \left\| \tilde{\boldsymbol{a}}_{\mathrm{r}} \right\| \left\| \tilde{\boldsymbol{a}}_{\mathrm{s}} \right\| + \epsilon } \right)^{2}. 
\label{eq_dpar_orth} 
\end{equation}
Although posterior anchors are not directly utilized as inputs for the routing computation, Eq.~(\ref{eq_dpar_orth}) implicitly regularizes the routing distribution through the dependencies established in Eq.~(\ref{eq_dpar_anchor}). Its gradients update both $\tilde{w}_{\mathrm{p},\mathrm{r}}$ and $\boldsymbol{e}_{\mathrm{p}}$, driving a differentiated allocation of patch units across different experts. As a result, the additive terms passed to RGC are diversified to better align the patterns.

\subsection{Reliability-Gated Composition}
Although the additive formulation provides inherent interpretability, it remains vulnerable to background noise. That is, task-irrelevant patches may still exert a substantial influence on the final prediction, undermining the reliability of the attribution. To mitigate this issue, RGC calibrates each patch--expert term via expert reliability and patch admission prior to aggregation, strictly preserving the additive decomposed structure.

The core mechanism of RGC is the formulation of the calibrated patch logit $\boldsymbol{\ell}_{\mathrm{p}}$. Rather than uniformly summing all expert outputs, $\boldsymbol{\ell}_{\mathrm{p}}$ selectively aggregates the expert class logits $\boldsymbol{o}_{\mathrm{p},\mathrm{r}}$, modulated by a sparse routing weight $w_{\mathrm{p},\mathrm{r}}$ and an expert reliability score $\kappa_{\mathrm{p},\mathrm{r}}$:
\begin{equation}
\boldsymbol{\ell}_{\mathrm{p}}
=
\sum_{\mathrm{r}=1}^{\mathrm{M}}
w_{\mathrm{p},\mathrm{r}}\kappa_{\mathrm{p},\mathrm{r}}\boldsymbol{o}_{\mathrm{p},\mathrm{r}} ,
\label{eq_rgc_patch_logit}
\end{equation}
where $\boldsymbol{o}_{\mathrm{p},\mathrm{r}}=\boldsymbol{W}_{\mathrm{cls}}\boldsymbol{z}_{\mathrm{p},\mathrm{r}}$ with $\boldsymbol{z}_{\mathrm{p},\mathrm{r}} = f_{\mathrm{r}}\!\left(\boldsymbol{e}_{\mathrm{p}}\right)$ denotes the representation extracted by the $\mathrm{r}$-th expert from the patch unit $\boldsymbol{e}_{\mathrm{p}}$ and $\boldsymbol{W}_{\mathrm{cls}}$ represents the shared classification head.

To construct the gating factors in Eq.~(\ref{eq_rgc_patch_logit}), RGC evaluates both expert confidence and routing sparsity. The reliability score $\kappa_{\mathrm{p},\mathrm{r}}\in[0,1]$ assesses the confidence of the generated representation $\boldsymbol{z}_{\mathrm{p},\mathrm{r}}$:
\begin{equation}
\kappa_{\mathrm{p},\mathrm{r}}
=
\sigma\!\left(\psi_{\kappa}^{\mathrm{r}}\!\left(\boldsymbol{z}_{\mathrm{p},\mathrm{r}}\right)\right),
\label{eq_rgc_reliability}
\end{equation}
where $\psi_{\kappa}^{\mathrm{r}}$ denotes the reliability head. Moreover, rather than utilizing the dense distribution $\tilde{w}_{\mathrm{p},\mathrm{r}}$ from Eq.~(\ref{eq_dpar_routing}), RGC derives the sparse routing weight $w_{\mathrm{p},\mathrm{r}}$ by restricting the composition to a subset $\mathcal{S}_{\mathrm{p}}$ containing the top-$K_{\mathrm{s}}$ experts:
\begin{equation}
w_{\mathrm{p},\mathrm{r}}
=
m_{\mathrm{p}}
\frac{
\mathbb{I}\!\left(\mathrm{r}\in\mathcal{S}_{\mathrm{p}}\right)
\tilde{w}_{\mathrm{p},\mathrm{r}}
}{
\sum_{\mathrm{j}\in\mathcal{S}_{\mathrm{p}}}\tilde{w}_{\mathrm{p},\mathrm{j}}
},
\label{eq_rgc_sparse_weight}
\end{equation}
where $\mathbb{I}(\cdot)$ indicates the indicator function. The sparse weights satisfy $\sum_{\mathrm{r}=1}^{\mathrm{M}} w_{\mathrm{p},\mathrm{r}}=m_{\mathrm{p}}$, so that a valid patch distributes a unit of weight over its top-$K_{\mathrm{s}}$ experts. This sparsity strictly confines non-zero contributions to specific patch-expert paths. 

Although the reliability score $\kappa_{\mathrm{p},\mathrm{r}}$ in Eq.~(\ref{eq_rgc_reliability}) suppresses unreliable expert signals at the patch-expert level, it does not guarantee the global relevance of the patch to the final prediction. Therefore, the routed reliabilities are aggregated into patch-level score $s_{\mathrm{p}}$, and normalized across all patches to form global admission weight $\pi_{\mathrm{p}}$: 
\begin{equation}
s_{\mathrm{p}}
=
\sum_{\mathrm{r}=1}^{\mathrm{M}}
w_{\mathrm{p},\mathrm{r}}\kappa_{\mathrm{p},\mathrm{r}},
\qquad
\pi_{\mathrm{p}}
=
\frac{m_{\mathrm{p}}s_{\mathrm{p}}}{\sum_{\mathrm{i}=1}^{\mathrm{P}} m_{\mathrm{i}}s_{\mathrm{i}}+\epsilon} ,
\label{eq_rgc_admission}
\end{equation}
where $m_{\mathrm{p}}$ serves as the validity mask. 
This formulation naturally suppresses $\pi_{\mathrm{p}}$ for patches with low overall reliability.

Finally, the global prediction $\boldsymbol{u}$ can be written as:
\begin{equation}
\boldsymbol{u}
=
\sum_{\mathrm{p}=1}^{\mathrm{P}}
\pi_{\mathrm{p}}\boldsymbol{\ell}_{\mathrm{p}}
=
\sum_{\mathrm{p}=1}^{\mathrm{P}}
\sum_{\mathrm{r}=1}^{\mathrm{M}}
\pi_{\mathrm{p}}w_{\mathrm{p},\mathrm{r}}\kappa_{\mathrm{p},\mathrm{r}}\boldsymbol{o}_{\mathrm{p},\mathrm{r}} ,
\label{eq_rgc_prediction}
\end{equation}
and its $\mathrm{c}$-th entry $u_{\mathrm{c}}$ directly satisfies the decomposition objective in Eq.~(\ref{eq_decomposition_goal}) with $A_{\mathrm{p},\mathrm{c}}=\pi_{\mathrm{p}}\ell_{\mathrm{p},\mathrm{c}}$. Ultimately, MVRE encodes the patches, DPAR diversifies the routing process, and RGC filters unreliable terms. This ensures that all contributions are directly readable from the forward pass while effectively suppressing noise.

\subsection{Model Training}
The additive logit $\boldsymbol{u}$ in Eq.~(\ref{eq_rgc_prediction}) aligns the prediction and attribution into a shared objective. Given the ground-truth label $y$, the overall training objective $\mathcal{L}$ is formulated as:
\begin{equation}
\mathcal{L}
=
\mathcal{L}_{\mathrm{CE}}
+
\lambda_{\mathrm{orth}}\mathcal{L}_{\mathrm{orth}}
+
\lambda_{\mathrm{ra}}\mathcal{L}_{\mathrm{ra}} ,
\label{eq_total_loss}
\end{equation}
where $\mathcal{L}_{\mathrm{CE}}$ is the classification loss formulated as the standard cross-entropy loss over the final prediction $\boldsymbol{u}$, $\mathcal{L}_{\mathrm{orth}}$ is the orthogonal loss previously defined in Eq.~(\ref{eq_dpar_orth}) to promote expert specialization, and $\mathcal{L}_{\mathrm{ra}}$ is the reliability alignment loss. The balancing factors $\lambda_{\mathrm{orth}}$ and $\lambda_{\mathrm{ra}}$ are introduced to trade off the contributions of the corresponding loss terms.

To formulate the alignment loss $\mathcal{L}_{\mathrm{ra}}$, the reliability gate $\kappa_{\mathrm{p},\mathrm{r}}$ is supervised using the actual class support of a patch. Crucially, to avoid trivial optimization solutions, this target should remain strictly decoupled from the gating mechanism itself. We therefore derive the target from the routing-only contribution, obtained by omitting $\kappa_{\mathrm{p},\mathrm{r}}$ from Eq.~(\ref{eq_rgc_patch_logit}):
\begin{equation}
\bar{\boldsymbol{\ell}}_{\mathrm{p}}
=
\sum_{\mathrm{r}=1}^{\mathrm{M}}
w_{\mathrm{p},\mathrm{r}}
\boldsymbol{o}_{\mathrm{p},\mathrm{r}} .
\label{eq_ra_free_logit}
\end{equation}
Let $\hat{y}=\operatorname*{arg\,max}_{\mathrm{c}}u_{\mathrm{c}}$ denote the predicted class. The support of the $\mathrm{p}$-th patch is the $\hat{y}$-th entry of $\bar{\boldsymbol{\ell}}_{\mathrm{p}}$, normalized across all patches: 
\begin{equation}
t_{\mathrm{p}}
=
\frac{m_{\mathrm{p}}\left|\bar{\ell}_{\mathrm{p},\hat{y}}\right|}
     {\max_{\mathrm{i}}m_{\mathrm{i}}\left|\bar{\ell}_{\mathrm{i},\hat{y}}\right|+\epsilon} .
\label{eq_ra_target}
\end{equation}
To align the reliability score $s_{\mathrm{p}}$ in Eq.~(\ref{eq_rgc_admission}) with this target, the following alignment loss is employed:
\begin{equation}
\mathcal{L}_{\mathrm{ra}}
=
\frac{1}{\mathrm{P}_{\mathrm{v}}}
\sum_{\mathrm{p}=1}^{\mathrm{P}}
m_{\mathrm{p}}\,
\rho\left(s_{\mathrm{p}},b_{\mathrm{p}}\right) ,
\label{eq_ra_loss}
\end{equation}
where $b_{\mathrm{p}}=\operatorname{sg}\left(t_{\mathrm{p}}\right)$ employs the stop-gradient operator, $\mathrm{P}_{\mathrm{v}}=\sum_{\mathrm{p}=1}^{\mathrm{P}}m_{\mathrm{p}}$ denotes the total number of valid patches, and $\rho(\cdot,\cdot)$ denotes the binary cross-entropy loss. Consequently, strongly supported patches obtain higher reliability scores, whereas background patches receive smaller admission weights $\pi_{\mathrm{p}}$.

Regarding computational efficiency, AnchorMoE scales linearly with the number of generated patches and avoids pairwise patch interactions. 
Detailed analysis can be found in Appendix~\ref{app:complexity}.

\section{Experiments}
In this section, we first outline the experimental settings, and then demonstrate experimental results with discussions. 
\subsection{Experimental Settings}

The evaluation is organized into four parts.
\begin{itemize}
    \item \textbf{Classification performance.} To evaluate predictive performance, we compare AnchorMoE against nine classifiers on 18 UEA benchmark datasets. 
    \item \textbf{Explanation evaluation.} To verify whether the explanations accurately reflect the model's underlying decision-making process, the proposed AnchorMoE is evaluated against post-hoc baselines on both real-world and synthetic datasets with known patterns. 
    \item \textbf{Ablation study.} To validate the necessity of each module, we perform ablation studies by systematically excluding key components of AnchorMoE and evaluating the degradation in both predictive and interpretability performance. 
    \item \textbf{Qualitative analysis.} To qualitatively demonstrate interpretability, patch-level evidence attribution for individual predictions is visualized, and the supplementary detailed expert anchor diversity analyses are provided in Appendix~\ref{app:diversity}.
\end{itemize}

\textbf{Datasets.} 
The 18 datasets from the UEA archive~\citep{bagnall2018uea} span diverse domains and vary significantly in scale, sequence length, and dimensionality. 
To quantitatively and qualitatively validate evidence localization, four synthetic datasets are specifically constructed. These datasets provide explicit ground-truth masks for class-defining segments, intentionally leaving distractors and context offsets unmarked. To maintain narrative conciseness and logical flow within the main text, comprehensive generation details are deferred to Appendix~\ref{app:faithfulness}.

\textbf{Compared methods.}
Nine multivariate time series classifiers serve as baselines, spanning Transformer-based models (PatchTST \cite{nie2023patchtst}, FEDformer \cite{zhou2022fedformer}), MLP and lightweight architectures (DLinear \cite{zeng2023dlinear}, LightTS \cite{zhang2022lightts}, TSLANet \cite{eldele2024tslanet}), multi-scale and convolutional models (MICN \cite{wang2023micn}, TimesNet \cite{wu2023timesnet}, MPTSNet \cite{mu2025mptsnet}), and a kernel-based model (MiniRocket \cite{dempster2021minirocket}). For explanation evaluation, the intrinsic attribution of AnchorMoE is compared against nine attribution baselines, including gradient-based methods (Vanilla Gradient \cite{simonyan2014saliency}, Input$\times$Gradient \cite{shrikumar2016inputgrad}, Integrated Gradients \cite{sundararajan2017axiomatic}, DeepLIFT \cite{shrikumar2017deeplift}, DeepSHAP \cite{lundberg2017shap}), perturbation- and learning-based explainers (DynaMask \cite{crabbe2021dynamask}, FIT \cite{tonekaboni2020fit}, TimeX \cite{queen2023timex}), and a Random baseline.

\textbf{Implementation details.}
Models are implemented in PyTorch and trained on a single NVIDIA RTX 4090 GPU. AnchorMoE utilizes AdamW, with hyperparameters (e.g., learning rate, batch size, patch dimensions, expert count) configured per dataset and documented in the provided configuration files. Training incorporates linear warmup for auxiliary losses, early stopping, and dropout regularization across the embedding, MVRE, routing, and expert modules. The official UEA train/test splits are strictly followed for evaluation. Unless otherwise stated, results are reported as the mean and standard deviation over random seeds.

\textbf{Evaluation metrics.}
Classification performance is assessed via accuracy (ACC) and macro F1-score (F1), with average rank summarizing relative algorithmic standing across datasets. Explanation evaluation follows two complementary settings: (1) on real-world datasets, feature retention is measured using the $I(20)$ metric, which follows the retain-and-retrain protocol of~\cite{queen2023timex} (with the masking-sensitivity considerations noted by~\cite{hooker2019roar}), and the Gap metric~\cite{samek2016evaluating}; (2) on synthetic datasets, evidence localization is evaluated via AUPRC, IoU@K, and Precision@K against ground-truth masks. Formal definitions of these faithfulness metrics are provided in Appendix~\ref{app:faithfulness}. Higher values indicate superior performance across all metrics. 


\begin{table*}[!t]
\centering
\scriptsize
\setlength{\tabcolsep}{3pt}
\renewcommand{\arraystretch}{0.6}
\caption{Classification ACC ($\uparrow$) and F1 ($\uparrow$), with the \colorbox{red!25}{best} and \colorbox{red!8}{second-best} results highlighted in different colors.}
\label{tab:comparison_all_metrics}
\resizebox{\textwidth}{!}{
\begin{tabular}{@{}lccccccccccc@{}}
\toprule
\multirow{2}{*}{Datasets} & \multirow{2}{*}{Metrics} & \multicolumn{10}{c}{Method} \\
\cmidrule(lr){3-12}
 & & PatchTST & FEDformer & DLinear & LightTS & MICN & TimesNet & TSLANet & MPTSNet & MiniRocket & AnchorMoE (ours) \\
\midrule
ArticularyWordRecognition & ACC & 0.978±0.002 & 0.752±0.126 & 0.973±0.003 & 0.973±0.003 & 0.954±0.014 & 0.978±0.003 & 0.976±0.066 & 0.977±0.000 & \cellcolor{red!8}{0.981±0.002} & \cellcolor{red!25}{0.992±0.004} \\
 & F1 & 0.978±0.002 & 0.735±0.133 & 0.973±0.003 & 0.973±0.003 & 0.952±0.014 & 0.978±0.003 & 0.911±0.078 & 0.977±0.000 & \cellcolor{red!8}{0.981±0.002} & \cellcolor{red!25}{0.992±0.004} \\
\midrule
BasicMotions & ACC & 0.700±0.000 & 0.850±0.020 & 0.875±0.000 & 0.975±0.020 & 0.925±0.020 & \cellcolor{red!8}{0.992±0.012} & \cellcolor{red!25}{1.000±0.000} & 0.950±0.025 & \cellcolor{red!25}{1.000±0.000} & \cellcolor{red!25}{1.000±0.000} \\
 & F1 & 0.695±0.005 & 0.845±0.023 & 0.873±0.002 & 0.975±0.021 & 0.925±0.020 & \cellcolor{red!8}{0.992±0.012} & \cellcolor{red!25}{1.000±0.000} & 0.950±0.024 & \cellcolor{red!25}{1.000±0.000} & \cellcolor{red!25}{1.000±0.000} \\
\midrule
CharacterTrajectories & ACC & 0.954±0.017 & 0.976±0.001 & 0.973±0.003 & 0.978±0.002 & \cellcolor{red!25}{0.999±0.007} & 0.987±0.002 & 0.979±0.001 & 0.985±0.001 & 0.916±0.003 & \cellcolor{red!8}{0.992±0.001} \\
 & F1 & 0.953±0.018 & 0.975±0.001 & 0.973±0.003 & 0.977±0.002 & \cellcolor{red!25}{0.997±0.007} & 0.985±0.002 & 0.977±0.002 & 0.984±0.001 & 0.911±0.003 & \cellcolor{red!8}{0.992±0.001} \\
\midrule
Cricket & ACC & 0.957±0.007 & 0.449±0.035 & 0.917±0.000 & 0.912±0.007 & 0.773±0.010 & 0.917±0.020 & 0.973±0.059 & 0.931±0.001 & \cellcolor{red!8}{0.986±0.000} & \cellcolor{red!25}{0.995±0.008} \\
 & F1 & 0.954±0.007 & 0.372±0.039 & 0.914±0.002 & 0.909±0.007 & 0.771±0.011 & 0.911±0.025 & 0.915±0.080 & 0.931±0.001 & \cellcolor{red!8}{0.986±0.000} & \cellcolor{red!25}{0.995±0.008} \\
\midrule
DuckDuckGeese & ACC & 0.260±0.033 & 0.287±0.009 & 0.627±0.009 & 0.513±0.009 & 0.465±0.015 & 0.567±0.019 & 0.519±0.047 & \cellcolor{red!8}{0.640±0.005} & 0.620±0.020 & \cellcolor{red!25}{0.670±0.012} \\
 & F1 & 0.220±0.042 & 0.243±0.012 & \cellcolor{red!8}{0.622±0.009} & 0.500±0.011 & 0.462±0.015 & 0.564±0.021 & 0.315±0.044 & 0.618±0.005 & 0.619±0.026 & \cellcolor{red!25}{0.673±0.017} \\
\midrule
FaceDetection & ACC & 0.667±0.006 & 0.665±0.003 & 0.680±0.002 & 0.666±0.004 & \cellcolor{red!8}{0.686±0.012} & 0.671±0.008 & 0.594±0.016 & 0.684±0.006 & 0.591±0.011 & \cellcolor{red!25}{0.696±0.007} \\
 & F1 & 0.664±0.008 & 0.665±0.003 & 0.680±0.002 & 0.666±0.004 & \cellcolor{red!8}{0.682±0.013} & 0.671±0.007 & 0.573±0.026 & 0.680±0.007 & 0.591±0.011 & \cellcolor{red!25}{0.693±0.011} \\
\midrule
FingerMovements & ACC & 0.583±0.026 & 0.573±0.009 & 0.590±0.029 & 0.597±0.029 & 0.594±0.012 & 0.577±0.005 & 0.525±0.030 & \cellcolor{red!8}{0.610±0.005} & 0.590±0.010 & \cellcolor{red!25}{0.627±0.012} \\
 & F1 & 0.578±0.030 & 0.567±0.007 & 0.588±0.030 & 0.592±0.032 & 0.591±0.012 & 0.574±0.004 & 0.499±0.053 & \cellcolor{red!8}{0.601±0.005} & 0.588±0.010 & \cellcolor{red!25}{0.622±0.015} \\
\midrule
Heartbeat & ACC & 0.709±0.005 & 0.745±0.005 & 0.756±0.004 & 0.756±0.004 & 0.770±0.011 & \cellcolor{red!8}{0.792±0.013} & 0.782±0.030 & 0.751±0.004 & 0.724±0.012 & \cellcolor{red!25}{0.810±0.008} \\
 & F1 & 0.454±0.019 & 0.628±0.007 & 0.627±0.016 & 0.617±0.025 & \cellcolor{red!8}{0.767±0.011} & 0.692±0.039 & 0.575±0.070 & 0.674±0.006 & 0.688±0.015 & \cellcolor{red!25}{0.797±0.015} \\
\midrule
InsectWingbeat & ACC & 0.459±0.015 & 0.572±0.009 & 0.190±0.001 & 0.647±0.001 & 0.583±0.012 & 0.595±0.008 & 0.565±0.004 & 0.597±0.005 & \cellcolor{red!25}{0.659±0.009} & \cellcolor{red!8}{0.651±0.011} \\
 & F1 & 0.455±0.015 & 0.569±0.009 & 0.182±0.002 & 0.645±0.002 & 0.580±0.012 & 0.592±0.009 & 0.192±0.004 & 0.569±0.005 & \cellcolor{red!25}{0.658±0.009} & \cellcolor{red!8}{0.648±0.011} \\
\midrule
JapaneseVowels & ACC & 0.956±0.003 & 0.975±0.005 & 0.968±0.002 & 0.951±0.008 & 0.781±0.011 & 0.972±0.005 & 0.974±0.005 & 0.966±0.020 & \cellcolor{red!8}{0.982±0.004} & \cellcolor{red!25}{0.986±0.002} \\
 & F1 & 0.952±0.004 & 0.972±0.006 & 0.966±0.002 & 0.951±0.007 & 0.779±0.011 & 0.973±0.006 & 0.971±0.005 & 0.963±0.020 & \cellcolor{red!8}{0.983±0.005} & \cellcolor{red!25}{0.988±0.002} \\
\midrule
NATOPS & ACC & 0.776±0.017 & 0.933±0.009 & \cellcolor{red!8}{0.943±0.003} & 0.939±0.008 & 0.725±0.010 & 0.850±0.012 & 0.925±0.062 & 0.931±0.107 & 0.922±0.015 & \cellcolor{red!25}{0.974±0.006} \\
 & F1 & 0.775±0.018 & 0.933±0.009 & \cellcolor{red!8}{0.943±0.003} & 0.936±0.008 & 0.722±0.010 & 0.848±0.013 & 0.898±0.085 & 0.927±0.108 & 0.922±0.015 & \cellcolor{red!25}{0.970±0.003} \\
\midrule
PEMS-SF & ACC & 0.865±0.007 & 0.852±0.007 & 0.823±0.020 & 0.891±0.009 & 0.864±0.009 & 0.850±0.030 & 0.730±0.179 & \cellcolor{red!25}{0.929±0.007} & 0.827±0.010 & \cellcolor{red!8}{0.907±0.004} \\
 & F1 & 0.857±0.006 & 0.845±0.007 & 0.816±0.021 & 0.887±0.009 & 0.860±0.009 & 0.846±0.030 & 0.669±0.202 & \cellcolor{red!25}{0.925±0.007} & 0.818±0.012 & \cellcolor{red!8}{0.899±0.005} \\
\midrule
PenDigits & ACC & 0.962±0.002 & 0.978±0.001 & 0.921±0.005 & \cellcolor{red!25}{0.991±0.006} & 0.977±0.006 & \cellcolor{red!8}{0.986±0.002} & 0.975±0.015 & 0.985±0.013 & 0.971±0.002 & \cellcolor{red!25}{0.991±0.001} \\
 & F1 & 0.962±0.002 & 0.979±0.001 & 0.921±0.004 & \cellcolor{red!8}{0.989±0.007} & 0.977±0.006 & 0.986±0.002 & 0.963±0.022 & 0.985±0.013 & 0.971±0.002 & \cellcolor{red!25}{0.991±0.001} \\
\midrule
RacketSports & ACC & 0.776±0.009 & 0.840±0.008 & 0.761±0.011 & 0.748±0.010 & 0.780±0.011 & 0.855±0.011 & 0.837±0.016 & 0.868±0.010 & \cellcolor{red!8}{0.871±0.004} & \cellcolor{red!25}{0.921±0.004} \\
 & F1 & 0.784±0.008 & 0.850±0.008 & 0.771±0.007 & 0.745±0.011 & 0.777±0.011 & 0.864±0.011 & 0.598±0.018 & 0.865±0.010 & \cellcolor{red!8}{0.878±0.003} & \cellcolor{red!25}{0.921±0.004} \\
\midrule
SelfRegulationSCP2 & ACC & 0.535±0.014 & 0.532±0.016 & 0.537±0.019 & 0.563±0.012 & 0.553±0.013 & 0.545±0.012 & 0.526±0.037 & \cellcolor{red!8}{0.564±0.059} & 0.539±0.010 & \cellcolor{red!25}{0.576±0.004} \\
 & F1 & 0.498±0.031 & 0.511±0.029 & 0.536±0.019 & 0.559±0.012 & 0.549±0.013 & 0.536±0.008 & 0.478±0.069 & \cellcolor{red!8}{0.561±0.060} & 0.538±0.009 & \cellcolor{red!25}{0.573±0.002} \\
\midrule
SpokenArabicDigits & ACC & 0.972±0.002 & 0.984±0.001 & 0.965±0.000 & \cellcolor{red!25}{0.999±0.006} & 0.982±0.000 & 0.991±0.001 & 0.981±0.006 & 0.979±0.119 & 0.988±0.001 & \cellcolor{red!8}{0.995±0.001} \\
 & F1 & 0.972±0.002 & 0.984±0.001 & 0.965±0.001 & \cellcolor{red!25}{0.996±0.006} & 0.982±0.000 & 0.991±0.001 & 0.978±0.006 & 0.975±0.120 & 0.988±0.001 & \cellcolor{red!8}{0.995±0.001} \\
\midrule
StandWalkJump & ACC & \cellcolor{red!25}{0.533±0.031} & 0.511±0.063 & 0.444±0.063 & 0.491±0.114 & \cellcolor{red!25}{0.533±0.030} & \cellcolor{red!25}{0.533±0.054} & 0.381±0.058 & \cellcolor{red!8}{0.523±0.031} & 0.400±0.067 & \cellcolor{red!25}{0.533±0.144} \\
 & F1 & 0.503±0.030 & 0.463±0.090 & 0.408±0.036 & 0.488±0.114 & 0.427±0.031 & 0.475±0.085 & 0.278±0.146 & \cellcolor{red!8}{0.519±0.030} & 0.330±0.067 & \cellcolor{red!25}{0.524±0.154} \\
\midrule
UWaveGestureLibrary & ACC & 0.857±0.004 & 0.508±0.089 & 0.821±0.002 & \cellcolor{red!25}{0.927±0.008} & 0.813±0.002 & 0.868±0.010 & 0.897±0.008 & 0.866±0.079 & \cellcolor{red!8}{0.918±0.007} & \cellcolor{red!25}{0.927±0.005} \\
 & F1 & 0.858±0.003 & 0.446±0.127 & 0.812±0.005 & \cellcolor{red!8}{0.923±0.008} & 0.812±0.001 & 0.867±0.011 & 0.894±0.009 & 0.863±0.080 & 0.917±0.006 & \cellcolor{red!25}{0.927±0.005} \\
\midrule
Avg.Rank (ACC) $\downarrow$ &  & 7.39 & 7.22 & 7.06 & 5.06 & 6.08 & 4.69 & 6.44 & 4.39 & 5.25 & 1.42 \\
Avg.Rank (F1) $\downarrow$ &  & 7.42 & 7.08 & 6.92 & 4.83 & 5.97 & 4.78 & 7.69 & 4.33 & 4.69 & 1.28 \\
\bottomrule
\end{tabular}
}
\end{table*}

\subsection{Classification Performance}
ACC, F1, and average rank of AnchorMoE against nine counterparts on the 18 multivariate datasets are reported in Table~\ref{tab:comparison_all_metrics}, with the best and second-best results in each row highlighted in different colors. 
The observations are as follows. 

\textbf{Consistent overall superiority:} 
AnchorMoE achieves the most robust overall ACC and F1 performance, ranking first by a substantial margin with average ranks of 1.42 and 1.28, respectively, compared to 4.39 and 4.33 for the next-best method, MPTSNet. This superiority is exceptionally consistent across the evaluation suite. By ranking first on 28 of the 36 dataset-metric entries and remaining within the top two across all instances, AnchorMoE demonstrates robust, dataset-agnostic efficacy. Furthermore, a critical-difference analysis (Appendix~\ref{sec:significance}) formally confirms this statistical superiority.

\textbf{Advantage on localized non-stationary signals:} 
Performance gains are most substantial on multivariate signals characterized by strong local structures, such as RacketSports and NATOPS. This advantage becomes particularly pronounced on the short, isolated events of Cricket, where AnchorMoE attains an accuracy of 0.995 relative to 0.449 for FEDformer. Such contrast empirically confirms that patch-level expert routing isolates localized discriminative evidence far more effectively than global frequency-domain decomposition paradigms.

\textbf{Boundary case analysis:} AnchorMoE ranks second under two specific conditions. First, on near-saturated datasets (e.g., CharacterTrajectories, SpokenArabicDigits), accuracies already exceed 0.99, rendering gaps of at most 0.007 practically negligible. Second, datasets lacking localized patterns attenuate our patch-level prior: InsectWingbeat's globally diffuse textures favor MiniRocket's dense kernels, while the globally correlated spatial structure of PEMS-SF aligns with MPTSNet's multi-scale periodic modeling. Rather than indicating fundamental flaws, these cases merely reaffirm AnchorMoE's exceptional suitability for prevalent MTSC tasks driven by sparse, localized evidence.

\begin{table*}[!t]
\centering
\caption{Faithfulness evaluation on seven real UEA datasets under $I(20)$ ($\uparrow$) and Gap ($\uparrow$) metrics.}
\label{tab:uea_interpretability_i20_gap}
\resizebox{\textwidth}{!}{%
\begin{tabular}{@{}llcccccccccc@{}}
\toprule
\multirow{2}{*}{\textbf{Dataset}} & \multirow{2}{*}{\textbf{Metric}} & \multicolumn{10}{c}{\textbf{Explanation Method}} \\
\cmidrule(lr){3-12}
 & & \textbf{Random} & \textbf{Vanilla} & \textbf{Input$\times$Grad} & \textbf{IG} & \textbf{DeepLIFT} & \textbf{DeepSHAP} & \textbf{DynaMask} & \textbf{FIT} & \textbf{TimeX} & \textbf{AnchorMoE} \\
\midrule
\multirow{2}{*}{\textbf{CharacterTrajectories}} & $I(20)$ & 0.771$\pm$0.003 & 0.819$\pm$0.005 & 0.851$\pm$0.002 & 0.920$\pm$0.003 & 0.851$\pm$0.002 & 0.891$\pm$0.002 & 0.871$\pm$0.006 & \cellcolor{red!25}{0.939$\pm$0.005} & 0.906$\pm$0.004 & \cellcolor{red!8}{0.938$\pm$0.002} \\
 & Gap & 0.007$\pm$0.013 & 0.228$\pm$0.000 & 0.277$\pm$0.000 & 0.333$\pm$0.000 & 0.277$\pm$0.000 & 0.333$\pm$0.012 & 0.216$\pm$0.000 & \cellcolor{red!8}{0.499$\pm$0.000} & 0.204$\pm$0.000 & \cellcolor{red!25}{0.620$\pm$0.000} \\
\midrule
\multirow{2}{*}{\textbf{SpokenArabicDigits}} & $I(20)$ & 0.748$\pm$0.007 & 0.887$\pm$0.005 & 0.909$\pm$0.005 & 0.926$\pm$0.003 & 0.909$\pm$0.005 & 0.920$\pm$0.007 & 0.888$\pm$0.003 & \cellcolor{red!8}{0.961$\pm$0.003} & 0.933$\pm$0.002 & \cellcolor{red!25}{0.962$\pm$0.001} \\
 & Gap & 0.058$\pm$0.016 & 0.115$\pm$0.000 & 0.179$\pm$0.000 & 0.252$\pm$0.000 & 0.179$\pm$0.000 & 0.279$\pm$0.014 & 0.217$\pm$0.000 & \cellcolor{red!8}{0.374$\pm$0.000} & 0.293$\pm$0.000 & \cellcolor{red!25}{0.476$\pm$0.000} \\
\midrule
\multirow{2}{*}{\textbf{Heartbeat}} & $I(20)$ & 0.693$\pm$0.037 & 0.649$\pm$0.018 & 0.696$\pm$0.020 & 0.675$\pm$0.015 & 0.696$\pm$0.020 & 0.699$\pm$0.020 & 0.724$\pm$0.017 & \cellcolor{red!25}{0.777$\pm$0.017} & 0.712$\pm$0.030 & \cellcolor{red!8}{0.729$\pm$0.015} \\
 & Gap & -0.008$\pm$0.011 & 0.070$\pm$0.000 & 0.109$\pm$0.000 & 0.110$\pm$0.000 & 0.109$\pm$0.000 & 0.128$\pm$0.019 & 0.129$\pm$0.000 & \cellcolor{red!8}{0.347$\pm$0.000} & 0.043$\pm$0.000 & \cellcolor{red!25}{0.431$\pm$0.000} \\
\midrule
\multirow{2}{*}{\textbf{NATOPS}} & $I(20)$ & 0.624$\pm$0.068 & 0.563$\pm$0.017 & 0.576$\pm$0.018 & 0.648$\pm$0.020 & 0.576$\pm$0.018 & 0.644$\pm$0.026 & 0.833$\pm$0.006 & \cellcolor{red!8}{0.844$\pm$0.006} & \cellcolor{red!25}{0.852$\pm$0.003} & 0.837$\pm$0.017 \\
 & Gap & 0.008$\pm$0.012 & -0.162$\pm$0.000 & -0.051$\pm$0.000 & -0.010$\pm$0.000 & -0.051$\pm$0.000 & 0.030$\pm$0.013 & 0.111$\pm$0.000 & \cellcolor{red!8}{0.238$\pm$0.000} & 0.080$\pm$0.000 & \cellcolor{red!25}{0.307$\pm$0.000} \\
\midrule
\multirow{2}{*}{\textbf{DuckDuckGeese}} & $I(20)$ & 0.353$\pm$0.012 & 0.320$\pm$0.053 & 0.327$\pm$0.023 & 0.320$\pm$0.035 & 0.327$\pm$0.023 & 0.307$\pm$0.050 & \cellcolor{red!25}{0.440$\pm$0.000} & \cellcolor{red!25}{0.440$\pm$0.035} & \cellcolor{red!8}{0.413$\pm$0.058} & \cellcolor{red!25}{0.440$\pm$0.020} \\
 & Gap & -0.004$\pm$0.010 & -0.033$\pm$0.000 & -0.036$\pm$0.000 & -0.033$\pm$0.000 & -0.036$\pm$0.000 & -0.034$\pm$0.003 & 0.080$\pm$0.000 & \cellcolor{red!8}{0.159$\pm$0.000} & 0.052$\pm$0.000 & \cellcolor{red!25}{0.192$\pm$0.000} \\
\midrule
\multirow{2}{*}{\textbf{PEMS-SF}} & $I(20)$ & 0.437$\pm$0.052 & 0.778$\pm$0.020 & 0.757$\pm$0.010 & 0.734$\pm$0.031 & 0.757$\pm$0.010 & 0.730$\pm$0.022 & 0.669$\pm$0.017 & \cellcolor{red!8}{0.790$\pm$0.035} & 0.586$\pm$0.029 & \cellcolor{red!25}{0.836$\pm$0.024} \\
 & Gap & 0.007$\pm$0.031 & 0.100$\pm$0.000 & 0.117$\pm$0.000 & 0.116$\pm$0.000 & 0.117$\pm$0.000 & 0.114$\pm$0.004 & 0.105$\pm$0.000 & \cellcolor{red!8}{0.148$\pm$0.000} & 0.082$\pm$0.000 & \cellcolor{red!25}{0.207$\pm$0.000} \\
\midrule
\multirow{2}{*}{\textbf{Cricket}} & $I(20)$ & 0.713$\pm$0.035 & 0.681$\pm$0.014 & 0.685$\pm$0.032 & 0.681$\pm$0.014 & 0.685$\pm$0.032 & 0.704$\pm$0.045 & 0.866$\pm$0.032 & \cellcolor{red!8}{0.880$\pm$0.021} & \cellcolor{red!8}{0.880$\pm$0.021} & \cellcolor{red!25}{0.940$\pm$0.008} \\
 & Gap & -0.008$\pm$0.011 & 0.061$\pm$0.000 & -0.006$\pm$0.000 & -0.007$\pm$0.000 & -0.006$\pm$0.000 & -0.004$\pm$0.005 & -0.003$\pm$0.000 & \cellcolor{red!8}{0.131$\pm$0.000} & 0.049$\pm$0.000 & \cellcolor{red!25}{0.186$\pm$0.000} \\
\bottomrule
\end{tabular}%
}
\end{table*}

\begin{table*}[!t]
\centering
\scriptsize
\setlength{\tabcolsep}{2.8pt}
\renewcommand{\arraystretch}{0.92}
\caption{Evidence localization on four synthetic datasets under AUPRC ($\uparrow$), IoU@K ($\uparrow$), and Precision@K ($\uparrow$).}
\label{tab:synthetic_evidence_localization}
\resizebox{\textwidth}{!}{
\begin{tabular}{@{}llcccccccccc@{}}
\toprule
\multirow{2}{*}{\textbf{Dataset}} & \multirow{2}{*}{\textbf{Metric}} & \multicolumn{10}{c}{\textbf{Explanation Method}} \\
\cmidrule(lr){3-12}
 & & \textbf{Random} & \textbf{Vanilla} & \textbf{Input$\times$Grad} & \textbf{IG} & \textbf{DeepLIFT} & \textbf{DeepSHAP} & \textbf{DynaMask} & \textbf{FIT} & \textbf{TimeX} & \textbf{AnchorMoE} \\
\midrule
\multirow{3}{*}{Localized-Context} & AUPRC & 0.199$\pm$0.009 & 0.148$\pm$0.010 & 0.165$\pm$0.015 & 0.256$\pm$0.083 & 0.165$\pm$0.015 & 0.241$\pm$0.050 & 0.207$\pm$0.016 & \cellcolor{red!8}{0.353$\pm$0.068} & 0.197$\pm$0.024 & \cellcolor{red!25}{0.584$\pm$0.015} \\
 & IoU@K & 0.122$\pm$0.009 & 0.056$\pm$0.006 & 0.093$\pm$0.005 & 0.188$\pm$0.094 & 0.093$\pm$0.005 & 0.186$\pm$0.076 & 0.155$\pm$0.042 & \cellcolor{red!8}{0.377$\pm$0.072} & 0.159$\pm$0.082 & \cellcolor{red!25}{0.434$\pm$0.015} \\
 & Precision@K & 0.190$\pm$0.014 & 0.086$\pm$0.008 & 0.134$\pm$0.012 & 0.250$\pm$0.107 & 0.134$\pm$0.012 & 0.251$\pm$0.087 & 0.217$\pm$0.051 & \cellcolor{red!8}{0.492$\pm$0.065} & 0.234$\pm$0.103 & \cellcolor{red!25}{0.560$\pm$0.025} \\
\midrule
\multirow{3}{*}{Composition-Context} & AUPRC & 0.389$\pm$0.013 & 0.317$\pm$0.021 & 0.348$\pm$0.031 & 0.458$\pm$0.110 & 0.348$\pm$0.031 & 0.457$\pm$0.073 & 0.411$\pm$0.016 & \cellcolor{red!8}{0.510$\pm$0.009} & 0.414$\pm$0.042 & \cellcolor{red!25}{0.643$\pm$0.011} \\
 & IoU@K & 0.250$\pm$0.007 & 0.175$\pm$0.026 & 0.217$\pm$0.037 & 0.322$\pm$0.092 & 0.217$\pm$0.037 & 0.306$\pm$0.056 & 0.272$\pm$0.029 & \cellcolor{red!8}{0.403$\pm$0.015} & 0.294$\pm$0.043 & \cellcolor{red!25}{0.444$\pm$0.008} \\
 & Precision@K & 0.384$\pm$0.009 & 0.279$\pm$0.038 & 0.335$\pm$0.048 & 0.457$\pm$0.106 & 0.335$\pm$0.048 & 0.441$\pm$0.063 & 0.406$\pm$0.037 & \cellcolor{red!8}{0.551$\pm$0.018} & 0.432$\pm$0.044 & \cellcolor{red!25}{0.579$\pm$0.009} \\
\midrule
\multirow{3}{*}{Distractor} & AUPRC & 0.392$\pm$0.018 & 0.512$\pm$0.024 & 0.573$\pm$0.022 & \cellcolor{red!8}{0.734$\pm$0.012} & 0.573$\pm$0.022 & 0.676$\pm$0.010 & 0.469$\pm$0.017 & 0.578$\pm$0.015 & 0.555$\pm$0.010 & \cellcolor{red!25}{0.799$\pm$0.018} \\
 & IoU@K & 0.250$\pm$0.007 & 0.409$\pm$0.020 & 0.461$\pm$0.022 & \cellcolor{red!8}{0.558$\pm$0.029} & 0.461$\pm$0.022 & 0.519$\pm$0.018 & 0.347$\pm$0.014 & 0.462$\pm$0.038 & 0.481$\pm$0.027 & \cellcolor{red!25}{0.580$\pm$0.013} \\
 & Precision@K & 0.384$\pm$0.009 & 0.559$\pm$0.016 & 0.605$\pm$0.020 & \cellcolor{red!8}{0.698$\pm$0.023} & 0.605$\pm$0.020 & 0.663$\pm$0.014 & 0.489$\pm$0.015 & 0.613$\pm$0.037 & 0.627$\pm$0.022 & \cellcolor{red!25}{0.719$\pm$0.012} \\
\midrule
\multirow{3}{*}{Multi-Distractor} & AUPRC & 0.396$\pm$0.005 & 0.492$\pm$0.037 & 0.552$\pm$0.033 & \cellcolor{red!8}{0.657$\pm$0.020} & 0.552$\pm$0.033 & 0.626$\pm$0.007 & 0.489$\pm$0.022 & 0.589$\pm$0.006 & 0.521$\pm$0.020 & \cellcolor{red!25}{0.746$\pm$0.006} \\
 & IoU@K & 0.253$\pm$0.006 & 0.385$\pm$0.023 & 0.436$\pm$0.035 & \cellcolor{red!8}{0.501$\pm$0.024} & 0.436$\pm$0.035 & 0.485$\pm$0.009 & 0.362$\pm$0.031 & 0.464$\pm$0.032 & 0.414$\pm$0.015 & \cellcolor{red!25}{0.513$\pm$0.009} \\
 & Precision@K & 0.387$\pm$0.010 & 0.531$\pm$0.023 & 0.581$\pm$0.035 & \cellcolor{red!8}{0.646$\pm$0.022} & 0.581$\pm$0.035 & 0.628$\pm$0.009 & 0.503$\pm$0.032 & 0.615$\pm$0.029 & 0.562$\pm$0.012 & \cellcolor{red!25}{0.662$\pm$0.006} \\
\bottomrule
\end{tabular}
}
\end{table*}

\begin{figure}[!t]
  \centering
  \includegraphics[width=\linewidth]{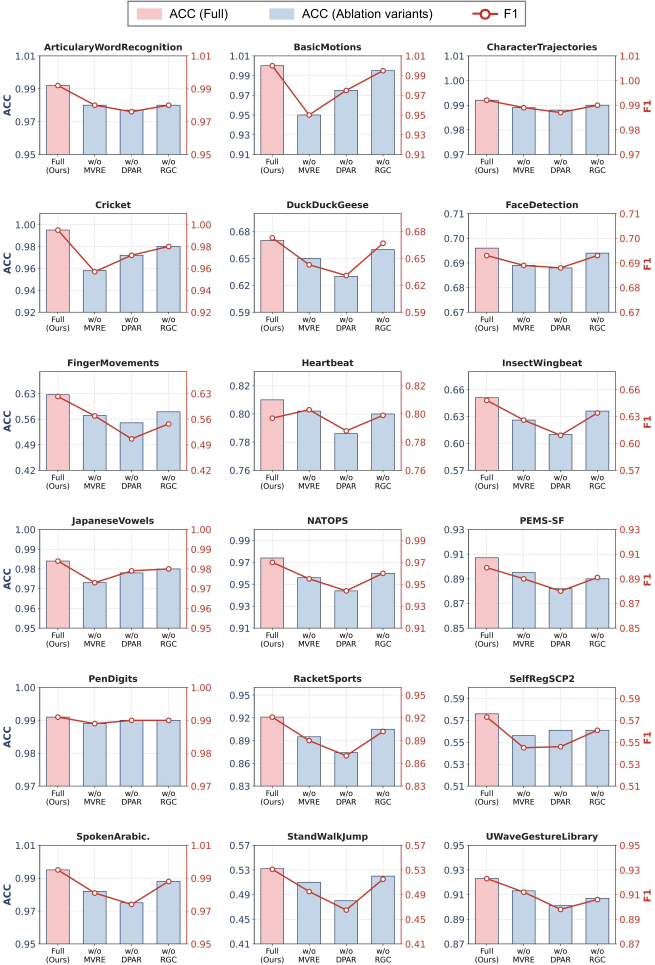}
  \caption{Component ablation on 18 UEA datasets.}
  \label{fig:ablation}
\end{figure}


\subsection{Explanation Evaluation}
\label{sec:explainability}
The faithfulness evaluation comprises feature retention testing on real-world UEA datasets alongside ground-truth mask alignment on synthetic benchmarks. AnchorMoE is benchmarked against nine post-hoc explainers to validate its intrinsic interpretability.

Real-world evaluations in Table~\ref{tab:uea_interpretability_i20_gap} expose a discrepancy in post-hoc baselines between subset informativeness ($I(20)$) and faithful importance ranking (Gap). For instance, while perturbation-based FIT occasionally matches or even surpasses AnchorMoE's $I(20)$ (e.g., on the \textit{Heartbeat}, \textit{CharacterTrajectories}, and \textit{DuckDuckGeese} datasets), it trails in Gap by margins of +0.03 to +0.12. Furthermore, a random baseline with near-zero Gap but a non-trivial $I(20)$ score confirms that $I(20)$ alone cannot establish overall faithfulness. In contrast, AnchorMoE consistently attains optimal Gap scores, verifying its faithful alignment with the decision process.

Synthetic evaluations in Table~\ref{tab:synthetic_evidence_localization} highlight severe performance inconsistencies among post-hoc baselines: basic gradient methods (e.g., Vanilla and Input$\times$Grad) fall to or even below the Random baseline in sparse and compositional settings (\textit{Localized-Context}, \textit{Composition-Context}), while perturbation methods like FIT fall to mid-rank against salient distractors (\textit{Distractor}). AnchorMoE inherently avoids these structural vulnerabilities, maintaining optimal metrics across all conditions. This robustness peaks exactly where baselines fail, outperforming the best competitor in AUPRC by 0.23 on \textit{Localized-Context}. Such stability derives from an inherent architectural synergy, i.e., the reliability gate suppresses high-variance patches prior to additive composition, while multi-view representations supply reliable evidence for accurate attribution.




\begin{figure}[!t]
  \centering
  \includegraphics[width=\linewidth]{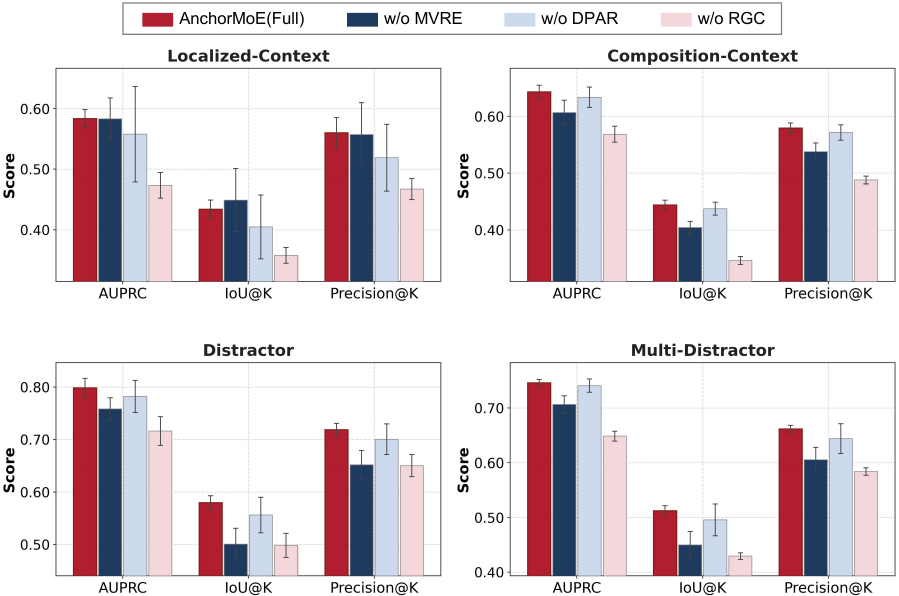}
  \caption{Evidence localization ablation on synthetic datasets.}
  \label{fig:interpretable_ablation}
\end{figure}

\subsection{Ablation Study}
\label{sec:ablation}
To quantify the contribution of individual mechanisms, we evaluate three reduced variants by ablating the MVRE, DPAR, and RGC components, respectively, across two evaluation axes: classification performance on real-world datasets (Figure~\ref{fig:ablation}) and evidence localization on synthetic benchmarks (Figure~\ref{fig:interpretable_ablation}).

Classification results demonstrate that the MVRE and DPAR components primarily drive discriminative capability. The DPAR component proves critical on challenging datasets such as \textit{FingerMovements}. Without the orthogonality constraint, experts collapse onto identical evidence, neutralizing the diversity essential for effective routing. Conversely, the MVRE component dominates in scenarios where single views are prone to misinterpretation, as observed on \textit{BasicMotions}. Although the RGC component also benefits classification, its impact on predictive accuracy remains moderate.

Evidence localization, however, reveals a reversed functional hierarchy where the RGC component becomes paramount. Without reliability weighting, background noise contaminates the additive decomposition, inducing severe degradation across all localization metrics. The MVRE component supports this process by supplying reliable upstream queries. Interestingly, DPAR exhibits limited direct influence on temporal boundaries. Its orthogonality constraint diversifies the captured feature types to resolve segment heterogeneity, rather than sharpening their precise temporal locations.

Collectively, these observations reveal an elegant architectural decoupling. That is, the DPAR component drives class discrimination via expert diversity, the RGC component ensures faithful evidence localization, and the MVRE component serves as the shared representational foundation for both objectives.

\begin{figure}[!t]
  \centering
  \includegraphics[width=\linewidth]{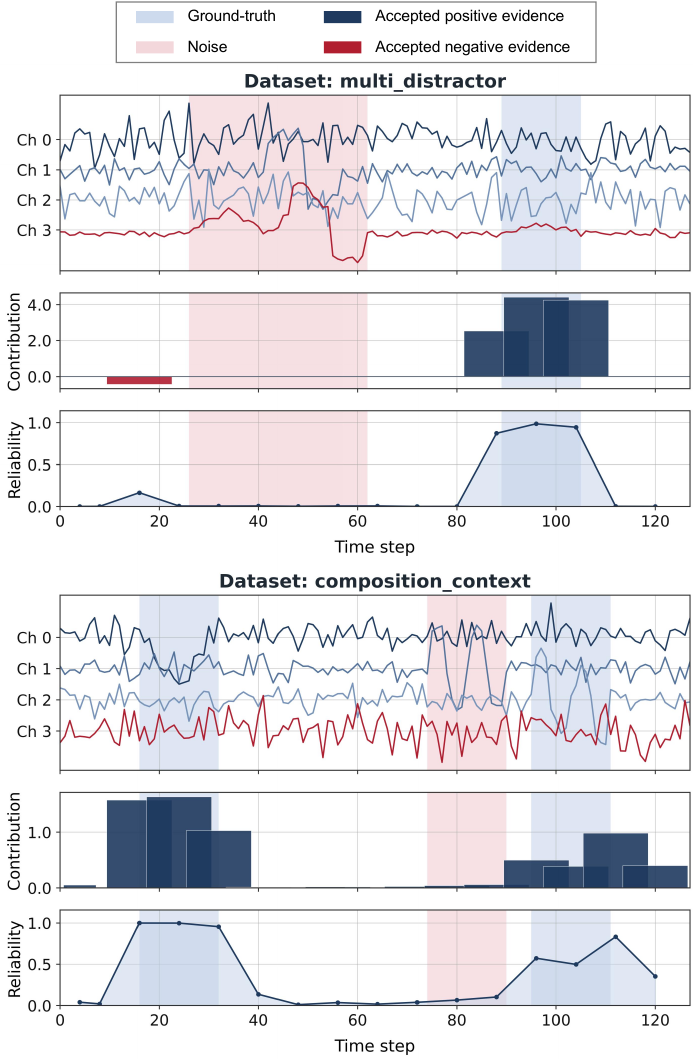}
  \caption{Visualization of evidence localization. Per-patch contributions and RGC reliability accurately localize within ground-truth evidence (blue) while attenuating across distractors and background (pink).}
  \label{fig:groundtruth}
\end{figure}

\subsection{Visual Results}
To qualitatively corroborate the quantitative faithfulness established previously, we visualize the decision-level evidence routing of AnchorMoE on synthetic datasets with annotated ground truth (i.e., \textit{Multi-Distractor} and \textit{Composition-Context} datasets). 

As illustrated in Figure~\ref{fig:groundtruth}, the routed patch contributions and the reliability scores generated by the RGC component align perfectly with the genuine evidence. The visualizations reveal that high reliability is exclusively assigned to regions carrying true discriminative features, strictly bypassing visually salient yet label-irrelevant distractors. Specifically, on \textit{Composition-Context} dataset, AnchorMoE successfully identifies and composes multiple distinct evidence fragments to form the final prediction, validating its local-evidence composition principle. On \textit{Multi-Distractor} dataset, it assigns decisive positive contributions to the ground-truth regions while effectively silencing the background noise. Even when minor negative background artifacts emerge, their associated reliability is profoundly suppressed, preventing them from influencing the final decision. Ultimately, these visual results intuitively confirm that the RGC component accurately binds isolated local patterns into trustworthy class evidence, robustly shielding the additive composition from background interference.

\section{Concluding Remarks}
This paper introduces AnchorMoE, an interpretable-by-design MoE framework for MTSC that inherently formulates predictions as exact additive decompositions over local patches. To robustly handle sparse signals and background noise, AnchorMoE integrates Multi-View Representation Embedding (MVRE) for rich evidence extraction, Diversified Posterior-Anchor Routing (DPAR) for non-redundant expert specialization, and Reliability-Gated Composition (RGC) to penalize uninformative segments. Evaluations confirm that AnchorMoE achieves highly competitive accuracy while preserving strict ante-hoc transparency. Although effective at localizing independent patterns, the additive formulation limits the capture of higher-order segment interactions, and the strict reliability gate may occasionally attenuate subtle contextual cues. Promising future directions include addressing these constraints and extending the transparent routing paradigm to broader domains like industrial anomaly detection and weather forecasting.



\begin{acks}
We gratefully acknowledge the creators and maintainers of the UEA multivariate time series classification archive~\cite{bagnall2018uea}, as well as the original donors of the individual datasets, for making these benchmarks publicly available. 
\end{acks}

\newpage

\bibliographystyle{ACM-Reference-Format}
\bibliography{references}

\appendix

\section{Complexity Analysis}
\label{app:complexity}
This section details the per-sample time complexity of AnchorMoE. Let $\mathrm{M}$ denote the number of experts, $\mathrm{d}$ the hidden dimension, $\mathrm{n}_{\mathrm{c}}$ the number of classes, $\mathrm{B}$ the number of frequency bands, $\mathrm{D}$ the number of variables, $\mathrm{L}$ the patch length, and $\mathrm{P}$ the number of generated patches in the patch grid. The validity mask $m_{\mathrm{p}}$ excludes padded patches from routing normalization, score composition, and loss computation, while the computation is constructed over the generated patch grid.

The total computational cost can be decomposed across the three primary modules. For MVRE, embedding each generated patch, transforming it into the spectral domain, and projecting it across the three views incurs $\mathcal{O}\left(\mathrm{P}\mathrm{D}\mathrm{L}\mathrm{d}\right)$, $\mathcal{O}\left(\mathrm{P}\mathrm{D}\mathrm{L}\log\mathrm{L}\right)$, and $\mathcal{O}\left(\mathrm{P}\left(\mathrm{D}\mathrm{B}\mathrm{d}+\mathrm{d}^{2}\right)\right)$ complexity, respectively, where the logarithmic factor arises from the fast Fourier transform along the patch length. In the DPAR module, mapping each patch query to a distribution over the $\mathrm{M}$ experts and aggregating the posterior anchors requires $\mathcal{O}\left(\mathrm{P}\mathrm{M}\mathrm{d}\right)$ operations. For RGC, evaluating all $\mathrm{M}$ experts and projecting their class logits prior to sparse masking dominates the per-patch overhead, requiring $\mathcal{O}\left(\mathrm{P}\mathrm{M}\left(\mathrm{d}^{2}+\mathrm{d}\mathrm{n}_{\mathrm{c}}\right)\right)$. Summing these contributions yields the overall per-sample complexity:
\begin{equation}
\mathcal{O}
\left(
\mathrm{P}\mathrm{D}\mathrm{L}\mathrm{d}
+
\mathrm{P}\mathrm{D}\mathrm{L}\log\mathrm{L}
+
\mathrm{P}
\left(
\mathrm{D}\mathrm{B}\mathrm{d}
+
\mathrm{d}^{2}
\right)
+
\mathrm{P}\mathrm{M}\mathrm{d}
+
\mathrm{P}\mathrm{M}
\left(
\mathrm{d}^{2}
+
\mathrm{d}\mathrm{n}_{\mathrm{c}}
\right)
\right) .
\label{eq_complexity}
\end{equation}

Given that the number of frequency bands satisfies $\mathrm{B}\le\mathrm{L}$ and the expert evaluation term $\mathrm{d}^{2}+\mathrm{d}\mathrm{n}_{\mathrm{c}}$ strictly dominates the routing term $\mathrm{d}$, Eq.~(\ref{eq_complexity}) simplifies to $\mathcal{O}\left(\mathrm{P}\mathrm{D}\mathrm{L}(\mathrm{d}+\log\mathrm{L})+\mathrm{P}\mathrm{D}\mathrm{B}\mathrm{d}+\mathrm{P}\mathrm{M}(\mathrm{d}^{2}+\mathrm{d}\mathrm{n}_{\mathrm{c}})\right)$. Here, the expert evaluation within RGC constitutes the leading cost.

This complexity formulation highlights two critical properties. First, every term is strictly linear with respect to the number of generated patches $\mathrm{P}$, entirely avoiding the $\mathcal{O}(\mathrm{P}^{2})$ bottleneck characteristic of standard attention-based encoders. Second, the calibrated terms $\pi_{\mathrm{p}}w_{\mathrm{p},\mathrm{r}}\kappa_{\mathrm{p},\mathrm{r}}o_{\mathrm{p},\mathrm{r},\mathrm{c}}$ that form the final attribution are explicitly computed during the forward pass. Consequently, extracting patch-level explanations reuses these intermediate quantities and requires no additional model forward or backward pass, introducing no additional asymptotic cost. AnchorMoE therefore achieves intrinsic interpretability without compromising scalability.

\section{Faithfulness Evaluation}
\label{app:faithfulness}
Faithfulness is evaluated on four synthetic multivariate datasets with ground-truth temporal evidence masks. Each sample has length $128$, four channels, and one of three classes. The base signal is Gaussian noise (std $0.18$), and class evidence is inserted as one or two motifs of length $16$ into class-dependent channels, drawn from sine, square, triangular, Gaussian-bump, and high-frequency sine patterns. The mask marks only the spans of class-defining motifs. The four datasets share this construction but differ in what competes with the true evidence. 

\paragraph{Localized-Context.} A single class-defining motif, plus a weak class-dependent offset spanning the last channel that provides context but is left unmarked. It tests whether a method identifies the compact motif rather than spreading importance onto the weak global context.

\paragraph{Composition-Context.} Two class-defining motifs, one in each half of the sequence, with the same unmarked global offset. It tests whether a method recovers multiple local units that jointly determine the label.

\paragraph{Distractor.} Two class-defining motifs and one strong but label-independent distractor, deliberately salient yet unmarked. It tests whether a method separates decision-relevant evidence from spurious saliency.

\paragraph{Multi-Distractor.} As above, but with three distractors, so that non-causal patterns are more frequent and prominent. It tests localization robustness under such interference.

\paragraph{Evaluation protocol.} We train on $900$ synthetic samples and evaluate attribution on $300$, restricted to correctly classified cases (up to $120$ per run), and report mean and standard deviation over three seeds.


\paragraph{Faithfulness metrics.} On real data, where the true evidence is unknown, we evaluate two retention metrics. $I(20)$ is the test accuracy of a shared evaluator retrained on the top 20\% most important time steps (with remaining steps masked). Gap measures the accuracy drop difference between most-relevant-first (MORF) and least-relevant-first (LERF) removals, with removed segments replaced by train-set means. On synthetic data, where the evidence is known, we report the Area Under the Precision-Recall Curve (AUPRC) over the full attribution ranking, alongside $\mathrm{IoU}@K$ and $\mathrm{Precision}@K$, where $K$ is the number of ground-truth patches. For all metrics, a higher value indicates better faithfulness.

\section{Significance Test}
\label{sec:significance}
\begin{figure}[!t]
  \centering
  \includegraphics[width=\linewidth]{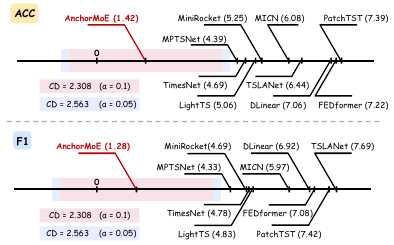}
  \caption{Critical difference diagram of the average ranks for 10 methods across 18 UEA datasets, evaluated on ACC (top) and F1 (bottom). AnchorMoE achieves the lowest average rank on both metrics. The shaded bands indicate the Bonferroni--Dunn critical difference relative to AnchorMoE at $\alpha=0.05$ ($\mathrm{CD}=2.563$) and $\alpha=0.10$ ($\mathrm{CD}=2.308$). All baselines fall outside these intervals.}
  \label{fig:cd}
\end{figure}

We statistically validate the performance improvements in Table~\ref{tab:comparison_all_metrics} via a critical difference (CD) analysis over the 18 UEA datasets~\cite{demvsar2006statistical}. 
\vfill\eject
As shown in Figure~\ref{fig:cd}, AnchorMoE significantly outperforms all nine competitors across the benchmark, with all baselines falling outside the critical difference threshold in both ACC and F1 panels. Specifically, the Friedman test rejects the null hypothesis of equal performance ($p<0.01$), with AnchorMoE achieving the lowest average rank on both ACC (1.42) and F1 (1.28). Given 10 methods, 18 datasets, and AnchorMoE as the control, the Bonferroni--Dunn critical difference is $\mathrm{CD}=2.563$ at $\alpha=0.05$. This critical value is derived from the two-sided Bonferroni-corrected normal quantile for $k-1$ comparisons against the control.

\section{Expert Diversity}
\label{app:diversity}

\begin{figure}[!t]
  \centering
  \includegraphics[width=\linewidth]{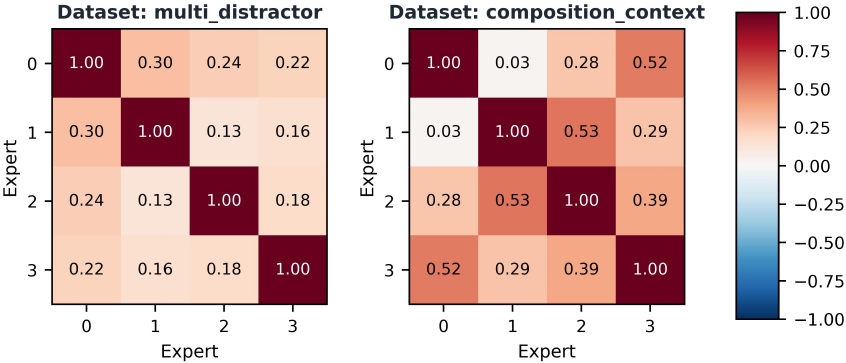}
  \caption{Pairwise cosine similarity matrix of posterior anchors for evaluating anchor orthogonality. Each posterior anchor corresponds to a specific expert, computed as the routing-weighted centroid of the evidence tokens assigned to it. The low off-diagonal values indicate reduced expert redundancy, suggesting that different experts capture distinct evidence representations.}
  \label{fig:diversity}
\end{figure}

To verify that DPAR mitigates expert redundancy, Figure~\ref{fig:diversity} visualizes the pairwise cosine similarity $S_{\mathrm{i},\mathrm{j}}=\cos(\tilde{\boldsymbol{a}}_{\mathrm{i}},\tilde{\boldsymbol{a}}_{\mathrm{j}})$ between posterior anchors. The off-diagonal values remain consistently low, indicating that the experts avoid feature redundancy and encode distinct evidence directions. The resulting orthogonal separation is prominent on the Multi-Distractor dataset but less pronounced on the Composition-Context dataset, where some anchor pairs exhibit moderate similarity. The observed similarity aligns with the structural characteristics of Composition-Context dataset, as evidence is explicitly shared across regions.

\end{document}